\documentclass[10pt,twocolumn,letterpaper]{article}

\usepackage{iccv}              
\usepackage{float}
\usepackage{mathtools}
\usepackage{stfloats} 

\usepackage[accsupp]{axessibility}
\definecolor{iccvblue}{rgb}{0.21,0.49,0.74}
\usepackage[pagebackref,breaklinks,colorlinks,allcolors=iccvblue]{hyperref}
\usepackage{multirow}
\def\paperID{*****} 
\def\confName{ICCV}
\def\confYear{2025}

\title{Bridging Domain Gaps for Fine-Grained Moth Classification Through Expert-Informed Adaptation and Foundation Model Priors}

\author{
Ross J. Gardiner\textsuperscript{1} \quad
Guillaume Mougeot\textsuperscript{2} \quad
Sareh Rowlands\textsuperscript{1} \\\quad
Benno I. Simmons\textsuperscript{1}\quad
Flemming Helsing\textsuperscript{2} \quad
Toke Thomas Høye\textsuperscript{2} \\
\textsuperscript{1}University of Exeter \quad
\textsuperscript{2}Aarhus University 
}

\begin{document}
\maketitle
\begin{abstract}
Labelling images of Lepidoptera (moths) from automated camera systems is vital for understanding insect declines. However, accurate species identification is challenging due to domain shifts between curated images and noisy field imagery. We propose a lightweight classification approach, combining limited expert-labelled field data with knowledge distillation from the high-performance BioCLIP2 foundation model into a ConvNeXt-tiny architecture. Experiments on 101 Danish moth species from AMI camera systems demonstrate that BioCLIP2 substantially outperforms other methods and that our distilled lightweight model achieves comparable accuracy with significantly reduced computational cost. These insights offer practical guidelines for the development of efficient insect monitoring systems and bridging domain gaps for fine-grained classification.

\end{abstract}    
\section{Introduction}
\label{sec:intro}

Insects are vital to terrestrial ecosystems and  global agriculture \cite{importance-of-insects}. They constitute roughly half of all animal species \cite{80percent-insects}, and terrestrial arthropods collectively account for 20 times the biomass of all wild mammals and birds combined \cite{biomass}. Despite their importance, insect populations are declining worldwide \cite{death-by-thousand-cuts, insect-decline-meta-analysis}, this remains poorly understood due to a lack of scalable field monitoring methods \cite{standard-insect-monitoring}.

In response, automated insect camera traps \cite{Insect-detect, amt, me-insectCT} autonomously photograph insects in-situ, enabling long-term and resource-efficient population monitoring \cite{standard-ai-insect}. But, the resulting raw imagery must be processed before it can provide actionable ecological insights. Advances in detection \cite{pollinator-detect,ecostack} and classification \cite{Hierarchical-insects, pucci-iet} have made computer vision the leading approach for this task \cite{deep-learning-will-transform-entomology}. 

Species-level recognition from camera-trap images remains difficult: morphological differences are subtle and expert taxonomists are scarce \cite{tax-imp}, making annotation expensive. Citizen science repositories such as the Global Biodiversity Information Facility (GBIF) \cite{gbif} provide labelled images for many taxa and are popular sources of training data. However, domain shift (differences in pose, lighting, and image quality between GBIF and automated camera trap imagery) reduces performance when models are trained exclusively on ``source-domain'' GBIF images and evaluated on ``target-domain'' in-situ insects \cite{jain}.

In addition, developing \emph{lightweight} computer vision models for insect monitoring remains important because: (1) ecologists worldwide may lack the computing infrastructure to run large models efficiently; (2) lightweight models are easier to retrain, which is valuable as most insect species remain undiscovered \cite{80percent-insects} and new species must be integrated; (3) lightweight architectures facilitate `edge' computing directly on camera trap hardware in remote environments. This is  common in insect camera trap designs \cite{darras, towards-edge, standard-ai-insect} where on-device inference is critical when connectivity constraints prohibit remote inference. Edge inference also allows the detection of specific species of interest on the device, allowing for the removal of unwanted images, conserving storage space for long deployments \cite{me-insectCT, motion-vectors}. 

We develop and evaluate a lightweight architecture that addresses domain shift through two strategies: (1) integrating expert-labelled target-domain data into training, and  (2) guiding learning via knowledge distillation (KD) from large-scale pretrained foundation models. We evaluate on a fine-grained and domain-shifted dataset of moth images from Automated Monitoring of Insects (AMI) cameras \cite{jain} deployed around 12 sites in Denmark. Combining expert-labelled target-domain data and KD enables our lightweight model to achieve comparable performance to BioCLIP and BioCLIP2, despite significantly lower parameter count. We provide recommendations for ecologists to use foundation models for their applications, and for the computer vision community towards fine-grained domain adaptation.

\begin{figure*}[htp]
\centering
    \includegraphics[width=0.8\textwidth]{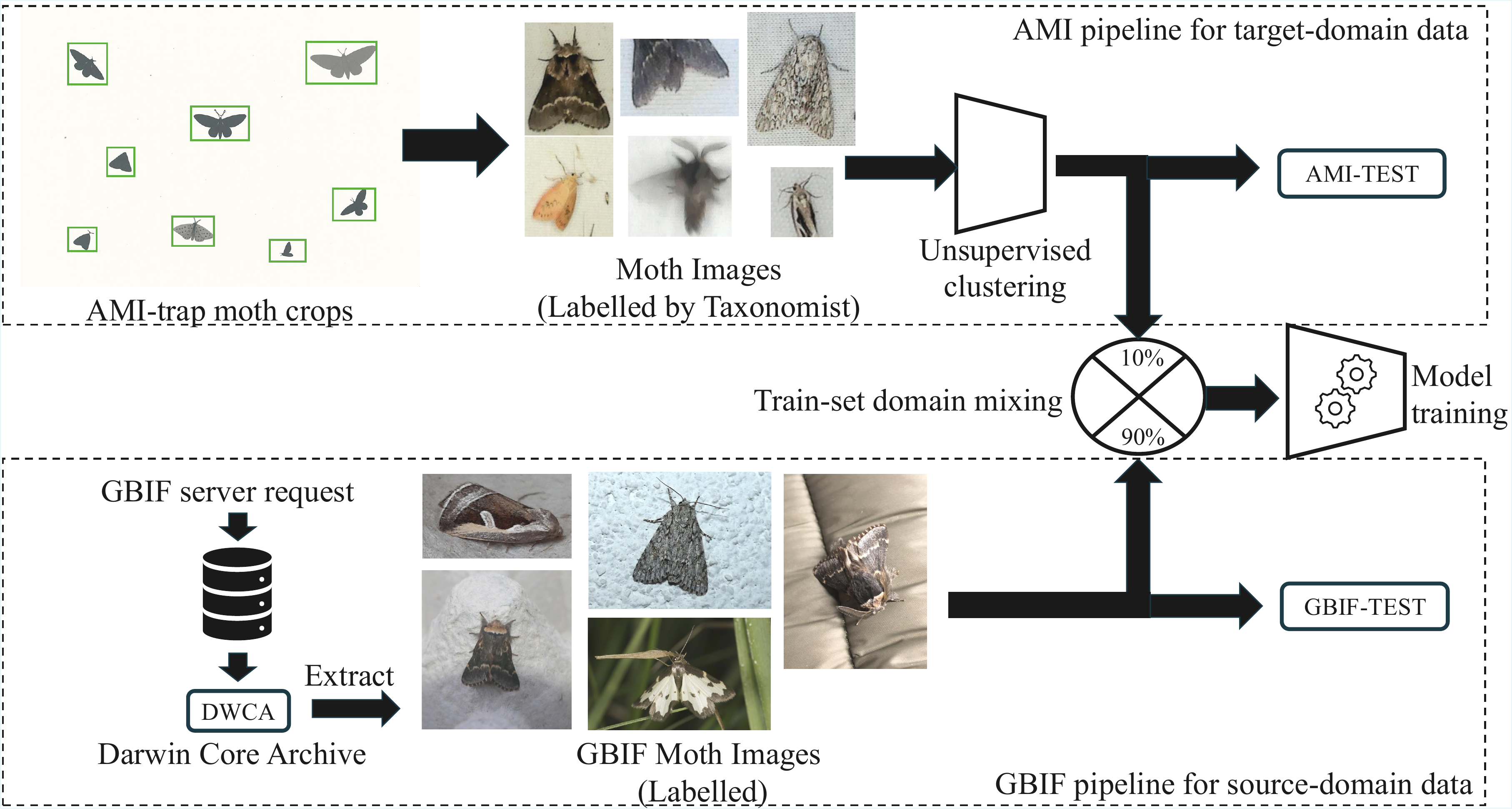}
    \caption{Data processing pipeline showing how moth images from AMI (target-domain) and GBIF (source-domain) are curated, labelled at the species level, and split into training and test sets. Controlled mixing of domains enables analysis of domain shift effects. Example images illustrate  differences between the two domains.}
    \label{fig:pipeline}
\end{figure*}

\subsection{Related Work}
There are many established methods for domain adaptation \cite{domain-adapt}, including feature alignment, adversarial learning and self-training. Synthetic data generation \cite{synthetic} has also previously been shown to improve generalisation for insect images. Here, we explore domain adaptation through supervised mixing of the training set as it is straightforward to implement and understand and allows us to leverage the limited expert-labelled target data available in our scenario. 

The use of CLIP-style \cite{clip} foundation models offers another route to cross-domain generalisation, pre-training large-scale representations transferable to diverse target domains. For biological images, recent models such as Insect-Foundation \cite{insectfoundation}, CLIBD \cite{clibd} and Biotrove \cite{biotrove} have shown strong performance on fine-grained insect identification tasks. Notably, BioCLIP \cite{bioclip}, trained on the TreeOfLife-10M dataset, and its successor BioCLIP2 \cite{bioclip2}, scaled to 214 million mostly GBIF-sourced images, achieve state-of-the-art accuracy. To retain the broad domain knowledge of such models while adapting to specific target tasks, we use KD, training smaller student model is to match the features of a larger pre-trained teacher \cite{kd-survey}. In this context, KD serves as a domain adaptation strategy, transferring generalisable representations from teacher to student \cite{dfkd}.

Our work is related to \cite{jain}, which introduced the AMI dataset and demonstrated ConvNeXt \cite{convnext} performance on cross-domain insect classification. We benchmark foundation models using a balanced dataset of Danish moth species and explore specialist model training with limited domain-specific data. Foundation models have also been used for camera trap classification; \cite{zero-shot-camera-trap} applied pre-trained models for zero-shot vertebrate identification, showing strong adaptability. We extend this by evaluating BioCLIP2 on AMI insect images, assessing the impact of expert-labelled target data and knowledge distillation for smaller models.

\section{Dataset}
We used a dataset of images of moths species collected from 12 AMI systems deployed in three regions of Denmark across three years (2022-2024) \cite{towards-edge}. Images captured by AMI are high resolution photographs of a large white screen, where moths are attracted to a UV light. Moths themselves make up small portions of each image and pre-existing object detection methods give crops for each moth instance \cite{standard-insect-monitoring}. Crops can be highly variable in size, lighting and quality, \Cref{fig:pipeline} shows some examples. And, following the long-tailed distribution typical of ecological scenarios, there are typically many images of the most common species, and relatively few of the others \cite{inat2018}. Identifiable moths in the AMI images have been labelled to the species level by one of the authors, an expert insect taxonomist specialised in moth species identification. As a primary focus in this study is domain shift, for our experiments, we use a subset of AMI moth crops represented by 101 species, where each species is represented by 110 AMI images, to isolate and study domain-shift effects. We refer to these throughout as the ``target-domain'' examples.

Individual images in the target-domain can be highly correlated, as data are obtained by time-lapse, and frequently an individual is stationary for long periods and is therefore imaged repeatedly in the same position. To prevent leakage of highly correlated images from the training set into our test set, we use an unsupervised deep clustering method to split the train/test set for each class (see Supplementary \Cref{app:cluster}). Of the 110 images per class, a test set of 10 is withheld, leaving 100 per class for training. 

To leverage GBIF images, we download each directly using an automated toolkit, supplying the GBIF species ID for each AMI class and filtering for the `imago' or adult life-stage, matching AMI life-stage. For each class, we obtained 224 images from unique GBIF occurrences, forming the ``source-domain" dataset. These are randomly split into 184 training and 20 withheld test images per class.

We progressively increase the proportion of target-domain images in training. At each step, we include as many GBIF images as possible to maximise training size. Once all available target images are used (over 35\%), further increases in the target proportion require reducing GBIF images. We constructed eight training sets with target-domain contributions of 0\%, 1\%, 5\%, 10\%, 20\%, 25\%, 33\%, and 50\% (statistics in Supplementary \Cref{app:stats}).

\section{Methodology}
\begin{figure}[htp]
\centering
    \includegraphics[width=0.8\linewidth]{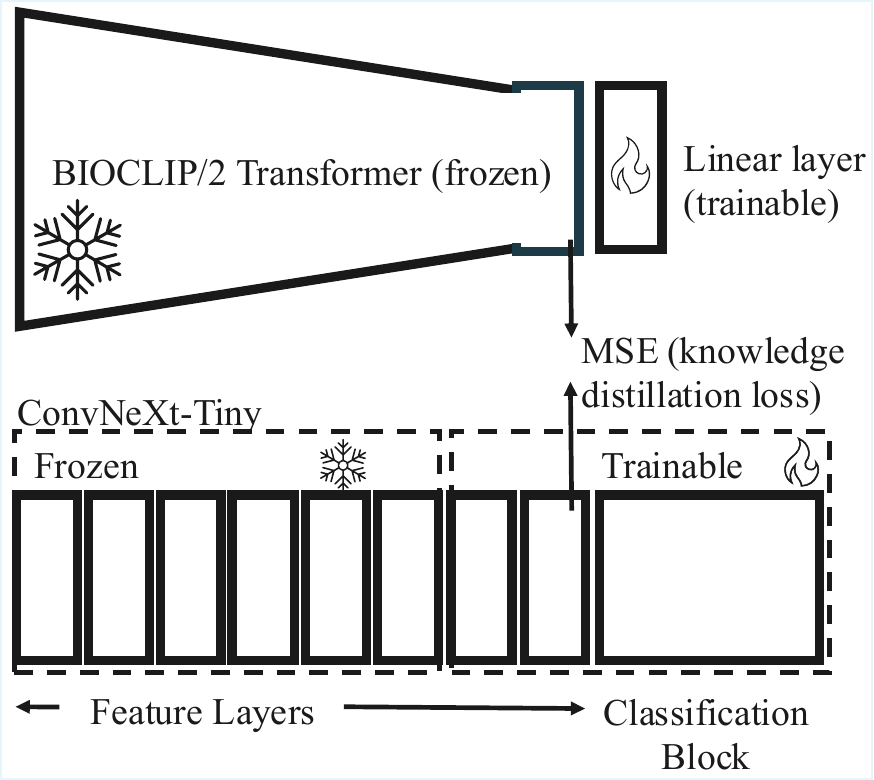}
    \caption{Foundation vision encoders (top) with a trainable classification linear layer. ConvNeXt instance (bottom) with pre-trained feature layers 1-6 frozen, the final feature layer is used for the mean-squared error (MSE) distillation loss for ConvNeXt+KD. }
    \label{fig:architecture}
\end{figure}

We train four models for moth species classification across domain-mixed datasets. To benchmark BioCLIP and BioCLIP2, we train linear classifiers on their frozen vision encoder outputs (as shown in \Cref{fig:architecture}). We also train a partially fine-tuned ConvNeXt-tiny model, unfreezing its classification head and top two layers, initialised with ImageNet-1K weights; this setup was found to perform best empirically. Finally, we apply KD from BioCLIP2 to ConvNeXt-tiny to improve generalisation through richer feature representations.

\subsection{Knowledge Distillation}

We use feature-based KD \cite{kd-survey}, where the student network aims to match the learned feature representations from the teacher to its own internal embeddings. Specifically, we implement a `hint' loss, defined as the mean squared error between the final feature representation of ConvNeXt and the output embedding of BioCLIP2, defined by \Cref{eqn:hint}:
 
\begin{equation}
    \mathcal{L}_{\text{hint}} = \frac{1}{N} \sum_{i=1}^{N} \left( s_i - t_i \right)^2
    \label{eqn:hint}
\end{equation}

Where $N = B \times C$, $B$ is the total number of elements in the batch and $C$ are the dimensions of the features, $s$ and $t$ represent the student and teacher embeddings, respectively. 

The hint loss is integrated into the ConvNeXt model loss using a weighting, defined by $\alpha$, to determine the contribution from categorical cross-entropy loss from the model output and the hint loss from features supervised by the teacher. Our loss function is described by \Cref{eqn:loss}:

\begin{equation}\label{eqn:loss}
       \mathcal{L}_{\text{total}} = \alpha \cdot \mathcal{L}_{\text{CE}}  + (1 - \alpha) \cdot \mathcal{L}_{\text{hint}}
\end{equation}

Where $\mathcal{L}_{\text{CE}}$ is the categorical cross-entropy loss. In our experiments, we set $\alpha$ equal to 0.5, assuming that classification and hint learning are equally important for our task. 

\subsection{Training}

All non-KD models are trained using categorical cross-entropy loss only. Each model is trained using an NVIDIA A100 GPU with the following hyperparameters: the learning rate, $\mu$, is set to $1e-3$, a weight decay of $1e-5$ is applied, the mini-batch size is 64 and the AdamW optimiser is used \cite{adamw}. Following \cite{jain}, we also deploy a MixRes strategy to augment the GBIF source-domain images only. For MixRes, given an image, it has a 25\% chance to be down-scaled to a size of 75x75 pixels or a 25\% chance to be downscaled to 150x150 pixels. Further augmentations are applied to all images though PyTorch \verb|RandAugment| function, with the \verb|num_ops| variable set to 2 and the \verb|magnitude| set to 3, each image is also 50\% likely to be horizontally flipped. All images are then resized to the model input size (224x224). 

\begin{table*}
  \centering
\begin{tabular}{ccccccccc}
\toprule
                                                                 & \multicolumn{2}{c}{\begin{tabular}[c]{@{}c@{}}ConvNeXt-tiny\\ (28M Params.)\end{tabular}}                                                   & \multicolumn{2}{c}{\begin{tabular}[c]{@{}c@{}}ConvNeXt-tiny+KD\\ (28M Params.)\end{tabular}}                                             & \multicolumn{2}{c}{\begin{tabular}[c]{@{}c@{}}BioCLIP\\ (86M Params.)\end{tabular}}                                                         & \multicolumn{2}{c}{\begin{tabular}[c]{@{}c@{}}BioCLIP2\\ (304M Params.)\end{tabular}}                                                       \\

\toprule
                                                                    \begin{tabular}[c]{@{}c@{}}Target-domain \\ Mix \\(\%)\end{tabular}      & \begin{tabular}[c]{@{}c@{}}Top-1 acc.\\ (target)\\ (\%)\end{tabular} & \begin{tabular}[c]{@{}c@{}}Top-1 acc.\\ (source)\\ (\%)\end{tabular} & \begin{tabular}[c]{@{}c@{}}Top-1 acc.\\ (target)\\ (\%)\end{tabular} & \begin{tabular}[c]{@{}c@{}}Top-1 acc.\\ (source)\\ (\%)\end{tabular} & \begin{tabular}[c]{@{}c@{}}Top-1 acc.\\ (target)\\ (\%)\end{tabular} & \begin{tabular}[c]{@{}c@{}}Top-1 acc.\\ (source)\\ (\%)\end{tabular} & \begin{tabular}[c]{@{}c@{}}Top-1 acc.\\ (target)\\ (\%)\end{tabular} & \begin{tabular}[c]{@{}c@{}}Top-1 acc.\\ (source)\\ (\%)\end{tabular} \\
\toprule
0\%                                                              & 59.4                                                                 & 88.1                                                                 & 64.7                                                                 & 91.2                                                                 & 71.2                                                                 & 95.2                                                                 & 88.3                                                                 & 97.6                                                                 \\
1\%                                                              & 60.4                                                                 & 86.9                                                                 & 63.0                                                                 & 90.4                                                                 & 74.4                                                                 & 95.1                                                                 & 87.4                                                                 & 98.3                                                                 \\
5\%                                                              & 72.8                                                                 & 87.1                                                                 & 79.4                                                                 & 90.9                                                                 & 76.6                                                                 & 95.0                                                                 & 89.4                                                                 & 98.4                                                                 \\
10\%                                                             & 77.7                                                                 & 87.6                                                                 & 81.0                                                                 & 90.6                                                                 & 78.0                                                                 & 95.1                                                                 & 91.5                                                                 & 98.5                                                                 \\
20\%                                                             & 83.0                                                                 & 87.9                                                                 & 85.0                                                                 & 90.8                                                                 & 81.1                                                                 & 95.0                                                                 & 90.0                                                                 & 98.0                                                                 \\
25\%                                                             & 82.5                                                                 & 87.1                                                                 & 86.6                                                                 & 91.0                                                                 & 82.3                                                                 & 94.9                                                                 & 91.5                                                                 & 98.0                                                                 \\
33\%                                                             & 85.9                                                                 & 88.1                                                                 & 86.4                                                                 & 89.9                                                                 & 83.9                                                                 & 95.1                                                                 & 91.3                                                                 & 98.0                                                                 \\
50\%                                                             & 85.4                                                                 & 84.5                                                                 & 89.4                                                                 & 88.8                                                                 & 85.8                                                                 & 94.4                                                                 & 91.6                                                                 & 97.8  \\
    \bottomrule

\end{tabular}
\caption{Showing top 1 source and target accuracy (acc.) percentages for each model architecture at each level of target-domain supervision. Model parameter counts (params) in millions (M) are also shown on the top row, beneath architecture names.}
  \label{tab:results}
\end{table*}

\section{Results}
\begin{figure} \includegraphics[width=0.9\linewidth]{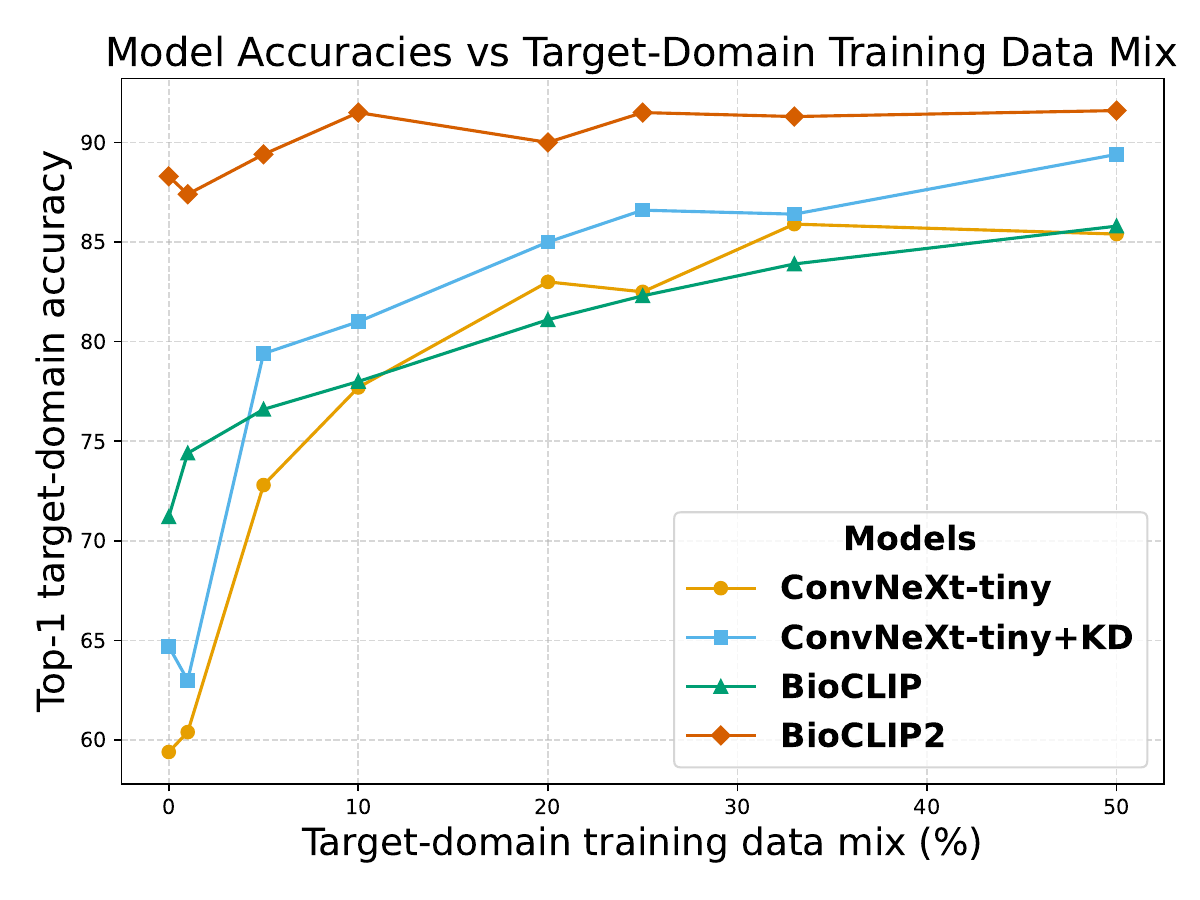}
    \caption{Top-1 target-domain accuracy plotted over training target-domain mix percentages for each architecture.}
    \label{fig:accuracies}
\end{figure}

For each dataset and architecture, we train for 10 epochs and report top-1 classification accuracy on the target and source test sets in \Cref{tab:results}. \Cref{fig:accuracies} shows target-domain accuracy; source-domain accuracy is discussed separately below.

BioCLIP2 consistently achieves the highest accuracy across all target-mix levels, particularly under low target supervision, with an average gain of 2.1\% over the range 1–50\% target data. ConvNeXt, ConvNeXt+KD, and BioCLIP show larger relative improvements with additional target data (average gains of 18.8\%, 16.8\%, and 9.1\%, respectively). ConvNeXt and ConvNeXt+KD perform poorly with minimal target supervision but match or exceed BioCLIP with more target supervision. Knowledge distillation provides a consistent boost across all target-mix levels, averaging +3.6\% over ConvNeXt.

Across most settings, source-domain accuracy remains higher than target for all models. For BioCLIP2, source accuracy is largely stable across target-mix levels. For the ConvNeXt variants and BioCLIP, there is a slight downward trend as more target data is included, reflecting the reduced amount of source-domain data in the mixed training set. The largest drop occurs at the 50\% mix. At this point, ConvNeXt and ConvNeXt+KD are the only cases where target accuracy marginally surpasses source accuracy.

\section{Discussion}

BioCLIP2 consistently achieves the highest performance, underscoring its strong generalisation capabilities and robustness to domain shift. This can be attributed to its pre-training on a very large and diverse image corpus, which likely exposed the model to a wide range of morphological traits, lighting conditions, and poses. Therefore, we recommend BioCLIP2 as a model with high transferability to AMI images. BioCLIP, while weaker overall, performs surprisingly well in low target-supervision settings, suggesting that its pre-trained features are also highly transferable to our target domain, but less so than BioCLIP2. 

Even modest target supervision  of 5\% significantly boosts target accuracy across all models, with particularly large gains of 13.4\% for ConvNeXt and 14.7\% for ConvNeXt+KD. The addition of KD from BioCLIP2 further enhances performance, suggesting it helps ConvNeXt learn deeper, target-relevant features beyond what target supervision provides. ConvNeXt+KD achieved the best accuracy per parameter, making it more suitable for compute-limited settings. Notably, at 50\% target supervision, it matched BioCLIP2’s performance without target supervision, despite having 10 times fewer parameters (28M vs. 304M). 

Overall, we provide two insights: (1) when the requirement for lightweight models is most important, we recommend KD and train-time domain mixing as effective strategies to build performant models; (2) in scenarios where ample computing power is available and/or target domain data are especially limited, we advocate using BioCLIP2 and saving precious expert-labelled data for model evaluation.

\section{Conclusion}
Through our experiments, we arrive at new directions for in-situ moth classification under domain shift. We show foundation models as accurate, adaptable classifiers and develop methods to exploit BioCLIP2 for training lightweight models. These advances are critical for the development of insect camera traps, which are a vital tool for understanding global insect declines. 

\section{Acknowledgements}
This article is based upon work from a short-term scientific mission (STSM) through the COST Action InsectAI, CA22129, supported by COST (European Cooperation in Science and Technology). This work was also funded via a doctoral training grant awarded as part of the UKRI AI Centre for Doctoral Training in Environmental Intelligence (UKRI grant number EP/S022074/1).





{
    \small
    \bibliographystyle{ieeenat_fullname}
    \bibliography{main}
}

\clearpage
\setcounter{page}{1}
\maketitlesupplementary

\section{Embedding-Based Clustering for Train/Test Splitting}\label{app:cluster}
\begin{figure}[!htp]
    \centering
    \includegraphics[width=\linewidth]{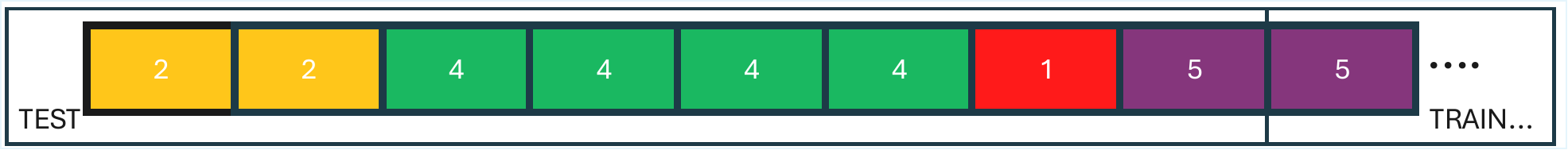}
    \caption{List of image embeddings with each showing its cluster number. Cluster groups have been shuffled. Each element are coloured and labelled by a unique cluster ID. The test set is partitioned as the first images in this list, with the remainder as train.}
    \label{fig:clustering}
\end{figure}

To obtain train/test splits from AMI data that are decoupled, we first embed each AMI image using a ResNet-50 network pretrained on ImageNet-1K; this places visually similar images near one another in the embedding space.
For every moth class, we then perform agglomerative hierarchical clustering with cosine distance, setting the number of clusters to $\max(K_{\text{min}}, \sqrt{N}\bigr)$, where $N$ is the number of images in the class and $K_{\text{min}}=5$.
After clustering, we randomly permute the cluster order and walk through this shuffled list, adding images to the test split until the specified amount of test examples are collected. This selective part is illustrated in \Cref{fig:clustering}. 

This approach enforces semantic separation between the train/test split, reducing bias and providing a realistic assessment of model generalisation.

\newpage
\section{Dataset Composition} \label{app:stats}

\label{sec:dataset}
\begin{table}[!htp]
\centering
\begin{tabular}{cccc}
\toprule
\begin{tabular}[c]{@{}c@{}}Target-domain \\ Mix \\ (\%)\end{tabular} & \begin{tabular}[c]{@{}c@{}}Target-domain\\ (AMI)\\ Contribution\end{tabular} & \begin{tabular}[c]{@{}c@{}}Source-domain\\ (GBIF) \\ Contribution\end{tabular} & \begin{tabular}[c]{@{}c@{}}Total\\ Dataset \\ Size\end{tabular} \\
\toprule
0\%                                                                  & 0                                                                            & 18573                                                                          & 18573                                                           \\
1\%                                                                  & 187                                                                          & 18573                                                                          & 18760                                                           \\
5\%                                                                  & 997                                                                          & 18573                                                                          & 19550                                                           \\
10\%                                                                 & 2063                                                                         & 18573                                                                          & 20636                                                           \\
20\%                                                                 & 4643                                                                         & 18573                                                                          & 23216                                                           \\
25\%                                                                 & 6191                                                                         & 18573                                                                          & 24764                                                           \\
33\%                                                                 & 9147                                                                         & 18573                                                                          & 27720                                                           \\
50\%                                                                 & 10100                                                                        & 10100                                                                          & 20200            \\
\bottomrule
\end{tabular}
\caption{Statistics of each domain-mixed training dataset at various amounts of target-domain percentages. Contributions from the target-domain (AMI) images and the source domain (GBIF) images vary to meet the desired domain mix fraction.}
\label{tab:stats}
\end{table}

\end{document}